\documentclass{sig-alternate-05-2015}

\begin{document}
\setcopyright{acmcopyright}
\doi{10.475/123_4}
\isbn{123-4567-24-567/08/06}
\conferenceinfo{PLDI '13}{June 16--19, 2013, Seattle, WA, USA}
\acmPrice{\$15.00}
\conferenceinfo{WOODSTOCK}{'97 El Paso, Texas USA}
\title{Improved Dense Trajectory with Cross Streams}
\numberofauthors{3} 

\author{
\alignauthor Katsunori Ohnishi\\
       \affaddr{Graduate School of Information}\\
       \affaddr{Science and Technology}\\
       \affaddr{University of Tokyo}\\
       \email{ohnishi@mi.t.u-tokyo.ac.jp}
\alignauthor Masatoshi Hidaka\\
       \affaddr{Graduate School of Information}\\
       \affaddr{Science and Technology}\\
       \affaddr{University of Tokyo}\\
       \email{hidaka@mi.t.u-tokyo.ac.jp}
\alignauthor Tatsuya Harada\\
       \affaddr{Graduate School of Information}\\
       \affaddr{Science and Technology}\\
       \affaddr{University of Tokyo}\\
       \email{harada@mi.t.u-tokyo.ac.jp}
}

\maketitle
\begin{abstract}
Improved dense trajectories (iDT) have shown great performance in action recognition, and their combination with the two-stream approach has achieved state-of-the-art performance. It is, however, difficult for iDT to completely remove background trajectories from video with camera shaking. Trajectories in less discriminative regions should be given modest weights in order to create more discriminative local descriptors for action recognition. In addition, the two-stream approach, which learns appearance and motion information separately, cannot focus on motion in important regions when extracting features from spatial convolutional layers of the appearance network, and vice versa. In order to address the above mentioned problems, we propose a new local descriptor that pools a new convolutional layer obtained from crossing two networks along iDT. This new descriptor is calculated by applying discriminative weights learned from one network to a convolutional layer of the other network. Our method has achieved state-of-the-art performance on ordinal action recognition datasets, 92.3\% on UCF101, and 66.2\% on HMDB51.
\end{abstract}
\def\vector#1{\mbox{\boldmath $#1$}}
\section{Introduction and Related Work}
\label{sec:intro}

\begin{figure}[t]
\begin{center}
   \includegraphics[width=0.98\linewidth]{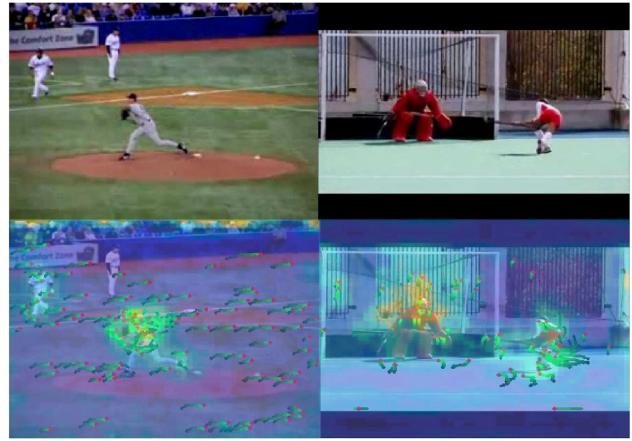}
\end{center}
\label{fig:visualized idt}
\vspace{-6mm} 
  \caption{Illustration of visualized iDT and feature map from convolutional layer in temporal net. We can see that there are some noisy trajectories in background due to camera shaking.}
\vspace{-4mm}
\end{figure}

Video representation is becoming increasingly important in today's online environment in which a massive amount of videos are uploaded on a daily basis. Various approaches have been proposed to efficiently and accurately represent the videos.

Dense trajectories \cite{wang2011action} and improved dense trajectories (iDT) \cite{wang2013action} have dominated action recognition. Extracting hand-crafted features \cite{dalal2005histograms, lucas1981iterative, dalal2006human} along these trajectories can provide effective local descriptors, and encoding these local descriptors with a Fisher vector (FV) \cite{perronnin2007fisher} or a vector of locally aggregated descriptors (VLAD) \cite{jegou2010aggregating} can provide an effective video representation \cite{kantorov2014efficient}.

Fueled by the recent success of convolutional neural networks (CNN) in image classification, video representations based on CNN have also been developed in action classification. The two-stream approach \cite{simonyan2014two} is one of the most successful methods that learns appearance information and motion information separately using one network whose input is RGB and the other network whose input is optical flow. The idea of this separate learning has been widely used in later works \cite{wang2015action, wu2015modeling, yue2015beyond}.

Aiming at fully end-to-end learning, three-dimensional CNN learning methods that can capture spatial and temporal information simultaneously and automatically \cite{sun2015human, tran2014c3d} have been developed recently. However, three-dimensional CNN learning is still a very difficult task, and these methods have not yet achieved comparable performance to the state-of-the-art approach.

Trajectory-pooled deep-convolutional descriptors (TDD) \cite{wang2015action} have shown state-of-the-art performance in action recognition by pooling convolutional two-stream layers along iDT. Because the convolutional layer retains position information, it is possible to combine it with iDT. However, TDD, which is based on iDT and the two-stream approach, has two main shortcomings: (1) as shown in Figure \ref{fig:visualized idt}, iDT cannot completely remove the background image for videos captured by a shaking camera. This can be solved by giving modest weights to background trajectories. (2) Although each network in the two-stream approach captures important information for action recognition, separate CNN learning sometimes lacks other important information that can be obtained only when spatial and temporal information are combined together. 

For example, thinking about the action of pitching a ball, it is difficult for spatial CNN to focus on the region around the pitcher's hand from only a single RGB image. This shortcoming makes it difficult to discriminate between similar actions such as the difference between a soccer penalty kick and a field-hockey penalty shot,  or between pitching a ball, a cricket shoot, a tennis serve, and a volleyball spike, especially when no background or context information is in the movie. Although iDT helps to solve this problem when recognizing action, iDT trajectories are hand-crafted so that they still contain both discriminative and non-discriminative trajectories. However, focusing on motion-important regions helps to extract more discriminative appearance features. When seeing a field hockey penalty, for example, the motion-important region is around the shooter's stick and the kipper. Extracting appearance features around these regions enables us to better recognize whether the player uses their own leg or the hockey stick or whether the kipper wears a protector or not. This can be also said in the case where spatial CNN and temporal CNN are reversed.

In order to address the above-mentioned problems, we utilize both networks in a two-stream approach by crossing two networks. Convolutional layers in spatial CNNs provide discriminative appearance features with position information while those in temporal CNNs provide discriminative motion features with position information. Thus, we propose a new descriptor that uses one network for the weights and gives these to the other network and pools along iDT, named cross-stream pooled descriptors (CPD). This is equivalent to pooling a convolutional layer of one network along iDT weighted by the convolutional layer of the other network, which leads to giving modest weights to iDTs in less discriminative regions. Our method has improved the performance of TDD on the ordinal action recognition datasets, UCF101\cite{soomro2012ucf101} and HMDB51\cite{kuehne2011hmdb}.

\section{Action Recognition Revisited}
In this section, we describe previous works on which our method is based.

\subsection{Improved Dense Trajectories}

Improved dense trajectories (iDT) \cite{wang2013action} are the improved version of dense trajectories \cite{wang2011action}, which can remove dense trajectories in background images considering camera motion. A video whose size is $(V_x, V_y, T)$ contains trajectories $P^k$ ($k=1\ldots K$):
\begin{equation}
\vspace{-1mm}
 P^k = \{(x^k_1, y^k_1, t^k_1), (x^k_2, y^k_2, t^k_2), \cdots , (x^k_L, y^k_L, t^k_L) \},
\end{equation}
where K is the number of trajectories in a video, $(x^k_l,y^k_l,t^k_l)$ is the position of the $l$th point in trajectory $P^k$, and $L$ is the length of trajectory. Following other works \cite{feichtenhofer2015dynamically, lan2015beyond, wang2013action, wang2015action}, we set $L=15$ in this paper.

\begin{figure*}[t]
\begin{center}
  	\includegraphics[width=0.9\linewidth]{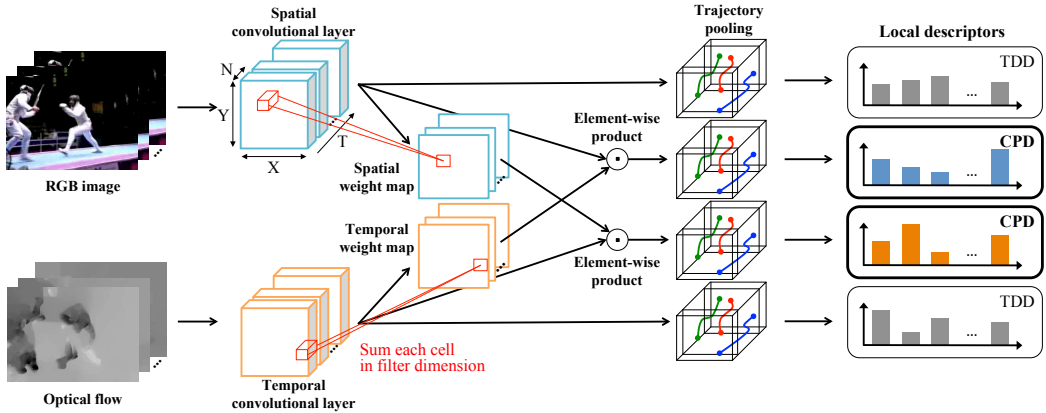}
\end{center}
	\vspace{-7mm}
   \caption{Illustration of the proposed local descriptors, named cross-stream pooled descriptors (CPD). After extraction, we encode these descriptors layer by layer and classify each of them. We then obtain final scores by simply summing all of their scores.}
	\vspace{-4mm}
   \label{fig:system}
\end{figure*}

\subsection{Two-Stream Approach}
The two-stream approach \cite{simonyan2014two} is a method that learns spatial information from RGB images and temporal information from optical flow images with each CNN separately. Since it is extremely difficult for a temporal net to learn motion using only single flow images, a sequence of ten frames are used as input. In this paper, we call the network learned from RGB images a `spatial network' and a network learned from optical flow images a `temporal network.'

\subsection{Trajectory-Pooled Deep-Convolutional Descriptors}
Trajectory-pooled deep-convolutional descriptors (TDD) \cite{wang2015action} combine iDT and the two-stream approach and achieves state-of-the-art performance on the UCF101 dataset. Given a ReLU applied convolutional layer $C \in \mathbb{R}^{X \times Y \times N \times T}$ from the two-stream approach, two normalization methods are applied to $C$, where $X$ and $Y$ are the width and height of the convolutional layer, $N$ is the number of channels, $T$ is the length of the video, and $C \geq 0$. Spatial normalization provides that $\widetilde{C}_{st}$ and channel normalization provides that $\widetilde{C}_{ch}$:\vspace{-1mm}
\begin{equation}
\label{equ:norm_st}
\widetilde{C}_{st}(x,y,n,t) = C(x,y,n,t)/{\rm max}_{x,y,t} C(x,y,n,t),
\vspace{-3mm}
\end{equation}
\begin{equation}
\label{equ:norm_ch}
\widetilde{C}_{ch}(x,y,n,t) = C(x,y,n,t) / {\rm max}_n C(x,y,n,t),
\vspace{-1mm}
\end{equation}
where $(x,y)$ is the position of the convolutional layer, n is the channel number of the convolutional layer, and t is the time in the video.

These $\widetilde{C}_{st}$ and $\widetilde{C}_{ch}$ are pooled along iDT instead of the originally pooled features (HOG \cite{dalal2005histograms}, HOF \cite{lucas1981iterative}, and MBH \cite{dalal2006human}). Given a normalized convolutional layer $\widetilde{C}^{a}_{b}$, which is the convolutional layer after applying spatiotemporal normalization or channel normalization ($b \in \{st, ch\}$) from spatial or temporal nets ($a \in \{sp, tmp\}$), proposed descriptors $TDD(P^k,\widetilde{C}^a_b) \in \mathbb{R}^{N}$ are obtained as follows:
\begin{equation}
\label{equ:tdd}
	TDD(P^{k},\widetilde{C}^{a}_{b}) = \sum^{L}_{l=1}\widetilde{C}^{a}( \overline{(r_x \times x^{k}_l)}, \overline{(r_y \times y^{k}_{l})} , t^{k}_{l}), 
\end{equation}
where $\overline{(\cdot)}$ is the rounding operation and $(r_x, r_y) = (X/V_w, Y/V_h)$. These descriptors are encoded by FV. The final video representation is obtained by concatenating encoded vectors from both normalization methods.

\section{iDT with the Cross Streams}
As described to this point, separate CNN learning cannot \\ always focus on truly important regions to capture an action's characteristics. Additionally, improved dense trajectories (iDT) cannot completely eliminate background trajectories from videos whose capturing camera experiences large motions. We address these problems to improve recognition performance using two equivalent methods. In this section, we describe both approaches in order to evaluate whether each problem can be improved by each calculation.

\subsection{Cross-Stream Pooling Along iDT}
In order to enhance motion-important regions in a spatial convolutional layer and appearance-important regions in a temporal convolutional layer, we propose a new convolutional layer for iDT pooling: the cross-stream layer. As shown in Figure \ref{fig:system}, we produce spatial and temporal convolutional layers element-wise and pool the resulting four-dimensional matrix along iDT. We call this method cross-stream pooled descriptors (CPD). However, since each of the {\it n}th filters in $C^{tmp}$ and $C^{sp}$ do not have the same meaning, the simple element-wise product ${C}^{tmp} (x,y,n,t) \times {C}^{sp}(x,y,n,t)$ might not work well. A convolutional layer shows large activation for discriminative regions. Thus, we can obtain a discriminative weight map $W \in \mathbb{R}^{X \times Y \times T}$ by simply taking the sum in the {\it n}-direction: 
\begin{equation}
\label{equ:summing}
	W^{tmp}(x,y,t) = \sum^{N}_{n=1} \widetilde{C}^{tmp} (x,y,n,t),
\end{equation}
where $\widetilde{C}^{tmp}$ is a normalized layer calculated from $C^{tmp}$ as in equations (\ref{equ:norm_st}) and (\ref{equ:norm_ch}). With this motion-based weight map, we can enhance the normalized spatial convolutional layer $\widetilde{C}^{sp}$, which contains appearance information:
\begin{equation}
	D^{sp} (x,y,n,t) = \widetilde{C}^{sp} (x,y,n,t) \times W^{tmp} (x,y,t).
\end{equation}
$D^{sp}$ represents new appearance features enhanced by motion-important regions.

Similarly to motion-based weights, we can obtain appearance-based weights $W^{sp}$ from $C^{sp}$, and $D^{tmp}$ is calculated in the same way. The term ‘cross stream’ originated from this cross utilization of two networks.

We then pool this $D$ along iDT as in equation (\ref{equ:tdd}) to obtain $CPD (P^{k}, D^{a}_{b}) \in \mathbb{R}^N $ as follows:
\begin{equation}
CPD (P^{k}, D^{a}_{b}) = \sum^{L}_{l=1} D^{a}_{b} ( \overline{(r_{x} \times x^{k}_l)}, \overline{(r_{y} \times y^{k}_{l})} , t^{k}_{l}).
\end{equation}

\subsection{Two-Stream Pooling Along Weighted iDT}
\label{sec:widt}
We next consider our method from a different point of view. Cross-stream pooled descriptors (CPD) can also be calculated as follows. 
In order to give modest weights to trajectories in the background region, we take advantage of the rest of the network. A convolutional layer $C^{tmp}$ obtained from a temporal CNN in the two-stream approach, for example, has discriminative motion features without losing position information. Using this $C^{tmp}$ as the weight and giving this weight to iDT, we can obtain new trajectories that are emphasized if they are in the region that contains motion-discriminative trajectories and are less emphasized if they are in regions that contain less motion-discriminative trajectories. As in equation (\ref{equ:summing}), we obtain a discriminative weight map $W^{tmp}$ by taking the sum in the $n$-direction. Each trajectory is weighted by this map $W^{tmp}$; then, we can obtain the weighted iDT. As for motion-based weights, an iDT weighted by an appearance-based map is calculated in the same way.
We then pool the normalized convolutional layer $\widetilde{C}^a$ along the emphasized iDT whose weights are calculated from $W^{\overline{a}}$ and obtain the CPD as follows:
\begin{equation}
	\begin{split}
	& CPD(P^{k},\widetilde{C}^{a}_{b}, W^{\overline{a}}_{b}) = \\
	 & \sum^{L}_{l=1}  W^{\overline{a}}_{b}(x^k_l,y^k_l,t^k_l) \times \widetilde{C}^{a}_{b}( \overline{(r_{x} \times x^{k}_l)}, \overline{(r_{y} \times y^{k}_{l})} , t^{k}_{l}). 
	 \end{split}
\end{equation}
This is equivalent to $CPD (P^{k}, D^{a}_{b})$.
\vspace{-2mm}

\section{Experiments}
\vspace{-3mm}

\begin{table}[tb]
\vspace{-2mm}
 \begin{center}
   \caption{Performance of each layer type on the UCF101 split1 dataset using parameters $(D, K) = (64, 128)$ for FV and $(D,K)=(128,64)$ for VLAD.}
 \small
  \begin{tabular}{|l|c|c|} 
\hline
Convolutional layer type			&FV		&VLAD		\\ \hline
(a) Spatial 					&81.2\%	&81.8\%		\\  
(b) Temporal 					&84.7\%	&85.5\%		\\ 
TDD: (a) + (b) 					&90.7\%	&91.5\%		\\ \hline
(c) Spatial weighted by temporal	&81.3\%	&82.9\%		\\ 
(d) Temporal weighted by spatial	&85.3\%	&85.9\%		\\ 
CPD ({\bf ours}): (c) + (d)			&90.4\%	&91.6\%		\\ \hline
TDD + CPD ({\bf ours})			&90.8\%	&92.0\%		\\  \hline
  \end{tabular}
 \label{tab:layer_type}
 \end{center}
 \vspace{-8mm}
\end{table}
\begin{table}[tb]
 \begin{center}
   \caption{The combination of convolutional layers resulting in each network on the UCF101 split1 dataset when VLAD is applied using parameters $(D,K)=(128,64)$. { (a), (b), (c), and (d) represent spatial, temporal, spatial weighted by temporal, and temporal weighted by spatial cases.}}
 \small
  \begin{tabular}{|l|c|c|c|c|} 
\hline
					&(a)		&(b)		&(c)		&(d)			\\\hline
Conv3				&71.9\%	&77.6\%	&74.1\%	&77.7\%		\\
Conv4				&78.2\%	&82.2\%	&78.5\%	&82.0\%	\\
Conv5				&76.3\%	&82.8\%	&75.7\%	&81.2\%	\\ \hline
Conv3 + Conv4			&79.0\%	&82.5\%	&80.3\%	&83.2\%	\\ 
Conv4 + Conv5			&81.3\%	&85.5\%	&81.5\%	&85.8\%	\\ \hline	
Conv3 + Conv4 + Conv5	&82.2\%	&85.8\%	&83.3\%	&86.5\%	\\ \hline
  \end{tabular}
 \label{tab:layer_comb}
 \end{center}
\vspace{-6mm}
\end{table}

\subsection{Datasets and Settings}
We conducted experiments on widely used action recognition datasets, UCF101 \cite{soomro2012ucf101} and HMDB51 \cite{kuehne2011hmdb}. We chose VGG16 \cite{Simonyan14c} as our CNN and utilized publicly available models \cite{wang2015towards} that had been already trained on UCF101. Because UCF101 has more variety of actions and videos, we used a model learned on UCF101 split 1 as the initial model for HMDB51 training. The learning rate and other training settings were the same as the training settings for UCF101\cite{wang2015towards}. We chose the models that showed the best validation scores during training.

As the convolutional layer for pooling, we chose conv3\_3, conv4\_3, and conv5\_3 from VGG16. We call these conv3, conv4, and conv5 in this paper, respectively. A final video representation of each layer was obtained by concatenating st-normed and ch-normed Fisher vectors following TDD\cite{wang2015action}. We fused SVM scores from each layer by taking the sum. Note that, in consideration of the calculation cost, we did not use multi-scale CNN, unlike TDD, and did not apply flipping or cropping to input images, unlike the original two-stream approach.

\vspace{-1mm}
\subsection{Analysis}
{\bf Parameters and Coding Methods}: We found the best coding method and parameters for TDD and CPD with UCF101 split1. Some previous works \cite{kantorov2014efficient,  xu2015discriminative} showed that VLAD encoding is also effective for action recognition. Thus, we tried both FV and VLAD for encoding. Through numerous experiments, we found that the best parameters for FV coding were $(D, K) = (64, 128)$, and those for VLAD coding were $(D,K)=(128,64)$, where $D$ is the dimension after compression by PCA and $K$ is the number of clusters. Details are given in the supplemental material owing to limited space here. 

{\bf Convolutional Type Combination:} Table \ref{tab:layer_type} shows that weighting the convolutional layer heightens accuracy for every layer and method, and combining our method with TDD improves the recognition accuracy of TDD. It is also shown that VLAD is more effective for all convolutional layer types than FV.

{\bf Layer Combination on Each Network:} Table \ref{tab:layer_comb} presents the combination patterns of convolutional layers in each network. In all network types, we can see that using all layers showed the best performance. Thus, we simply employed all of them.

\vspace{-1mm}
\subsection{Evaluation of CPD}

\begin{table}[tb]
\vspace{-8mm}
 \begin{center}
 \small
  \caption{Mean accuracy of CPD and other baseline methods on HMDB51 and UCF101. The score${\rm ^{*1}}$ of two-stream (VGG16) on HMDB51 in our calculation.}
  \vspace{0mm}
{\tabcolsep = 0.7mm
  \begin{tabular}{llllll||c|c|}
\hline
\multicolumn{6}{|l|}{Algorithm}											&HMDB51	&UCF101 		\\ \hline 
\multicolumn{6}{|l|}{iDT \& FV \cite{wang2013action}	}						&57.2\%		&85.9\%		\\
\multicolumn{6}{|l|}{Two stream \cite{simonyan2014two}}						&59.4\%		&88.0\% 		\\ 
\multicolumn{6}{|l|}{TDD \& FV \cite{wang2015action}}							&63.2\%		&90.3\%		\\ \hline  
\multicolumn{6}{|l|}{Two stream (VGG16)}									&${\rm 61.9\%^{*1}}$		&91.4\% \cite{wang2015towards}	\\ \hline
\multicolumn{3}{|l}{Spatial net } 		&\multicolumn{3}{l|}{(VGG16 w/o flip\&crop)}	&39.7\%		&75.5\%		\\ 
\multicolumn{3}{|l}{Temporal net} 	&\multicolumn{3}{l|}{(VGG16 w/o flip\&crop)}	&53.6\%		&81.0\%		\\ 
\multicolumn{3}{|l}{Two stream} 		&\multicolumn{3}{l|}{(VGG16 w/o flip\&crop)}	&59.3\%		&87.6\%		\\  \hline \hline
\multicolumn{6}{|l|}{TDD (VGG16) \& FV}									&63.2\%		&91.3\%			\\ 
\multicolumn{6}{|l|}{TDD (VGG16) \& VLAD}								&65.0\%		&92.0\%			\\ 
\multicolumn{6}{|l|}{CPD \& VLAD ({\bf ours})}								&65.2\%		&91.8\%			\\ \hline 
 \multicolumn{2}{|l}{TDD (VGG16)} 	& \multicolumn{2}{c}{ \multirow{2}{*}{+}}& \multicolumn{2}{l|}{CPD ({\bf ours})} & \multirow{2}{*}{\bf 66.2\%}	& \multirow{2}{*}{\bf 92.3\%} \\
  \multicolumn{2}{|l}{\& VLAD}		& &	& \multicolumn{2}{l|}{\& VLAD}	&&	\\
\hline
  \end{tabular}
  }
 \label{tab:cpd_ordinal}
\vspace{-8mm}
 \end{center}
\end{table}

\begin{figure}[t]
\vspace{-5mm}
\begin{center}
   \includegraphics[width=\linewidth]{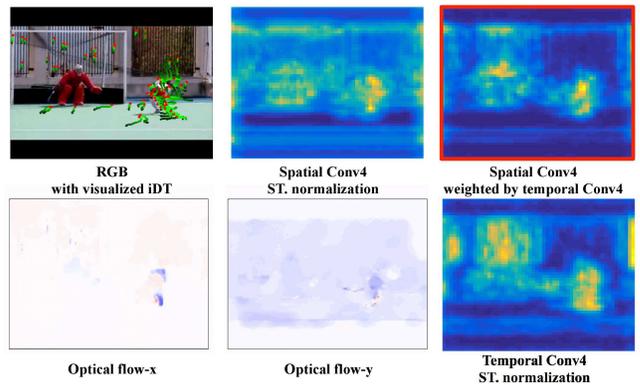}
\end{center}
\vspace{-8mm}
\caption{Sum of filter activations. We only show spatial layer weighted by temporal, because it will appear as the same image as temporal layer weighted by spatial with this visualization method.}
   \label{fig:visualize_ordinal}
\vspace{-5mm}
\end{figure}

Table \ref{tab:cpd_ordinal} represents the action accuracy of CPD and related methods on UCF101 \cite{soomro2012ucf101} and HMDB51 \cite{kuehne2011hmdb}, which are widely used action recognition datasets. Note that we did not flip and crop input images when predicting, unlike the original TDD. Although the two-stream approach of VGG16 without flipping and cropping shows worse performance than that of the original two-stream approach, as denoted in Table \ref{tab:cpd_ordinal}, the performance of TDD with FV is improved by replacing the CNN with VGG16. Encoding VLAD instead of FV also improves recognition accuracy. We then combine the scores of this TDD using VLAD with those of CPD, which increases the performance of TDD both on UCF101 and HMDB51.

Fig. \ref{fig:visualize_ordinal} shows an example of the visualized iDTs and convolutional layer activation. We can see that the spatial convolutional layer shows activation on many other objects whilst the convolutional layer weighted by the temporal convolutional layer shows activation mainly of the players. It can also be seen that some background iDTs still remain in the image due to camera shaking. However, the spatial layer weighted by the temporal layer activates mainly over the shooter and the kipper, ignoring their backgrounds. Thus, we can confirm that our method extracts appearance information mainly from motion-important regions and that these features capture different characteristics from those of TDD, which augments recognition performance.

\vspace{-1mm}
\subsection{Comparison with state-of-the-art}

\begin{table}[t]
 \begin{center}
 \small
 \vspace{2mm}
  \caption{Comparison with the state-of-the-art methods. The scores written inside of () is the accuracy when combined with iDT \& FV \cite{wang2013action}.}
 \vspace{-2mm}
{\tabcolsep = 0.4mm
  \begin{tabular}{|lc|lc|}
\hline
\multicolumn{2}{|c|}{HMDB51}							&\multicolumn{2}{c|}{UCF101} 					\\ \hline \hline 
iDT \& FV \cite{wang2013action}		&57.2\%			&iDT \& FV \cite{wang2013action}			&85.9\%		\\
iDT \& Stacked FV \cite{peng2014action}	&56.2\% 			&C3D \cite{tran2014c3d}					&85.2\%		\\
 \ \ + iDT \& FV 						&{\it (66.8\%)}		& \ \ + iDT \& FV 						&{\it (90.4\%)}		\\
${\rm F_{ST} CN}$ \cite{sun2015human}	&59.1\%			&${\rm F_{ST} CN}$	\cite{sun2015human}	&88.1\% 		\\
LATE \cite{feichtenhofer2015dynamically}	&62.2\%			&MIFS \cite{lan2015beyond}				&89.1\%			\\ 
TDD \& FV \cite{wang2015action}		&63.2\%			& TDD \& FV \cite{wang2015action}			&90.3\%	\\ 
 \ \ + iDT \& FV 						&{\it (65.9\%)}		& \ \ + iDT \& FV						&{\it (91.5\%)} \\
Video darwin \cite{fernando2015modeling}&63.7\%			&Hybrid LSTM \cite{wu2015modeling}		&91.3\%\\
MIFS \cite{lan2015beyond}			&65.1\%			&Two stream (VGG16) \cite{wang2015towards}	&91.4\%	\\ \hline \hline
CPD	({\bf ours})						&65.2\%			&CPD ({\bf ours})						&91.8\%		\\
TDD + CPD ({\bf ours})				&{\bf 66.2\%}		&TDD + CPD ({\bf ours})					&{\bf 92.3\%}	\\ 
\hline
  \end{tabular}
  }
 \label{tab:sota}
 \end{center}
\vspace{-6mm}
\end{table}

Table \ref{tab:sota} shows the comparison of our method with other methods of action recognition on the UCF101 and HMDB51 datasets. On UCF101, the proposed method achieved state-of-the-art performance: 0.8\% improvement over the combination of TDD \cite{wang2015action} and iDT \cite{wang2013action}. On HDMB51, our method achieved comparable performance to state-of-the-art methods. Considering the scores without adding iDT \& FV \cite{wang2013action}, our method shows the best performance.

\vspace{-3mm}
\section{Conclusion}
This study proposed a new type of local descriptors for action recognition, termed cross-stream pooled descriptors (CPD), that pools crossed convolutional layers along iDT. Our method achieved state-of-the-art performance on the widely used action recognition datasets UCF101 and HMDB51.

\bibliographystyle{abbrv}
\bibliography{paper}

\begin{thebibliography}{10}

\bibitem{dalal2005histograms}
N.~Dalal and B.~Triggs.
\newblock Histograms of oriented gradients for human detection.
\newblock In {\em CVPR}, 2005.

\bibitem{dalal2006human}
N.~Dalal, B.~Triggs, and C.~Schmid.
\newblock Human detection using oriented histograms of flow and appearance.
\newblock In {\em ECCV}, 2006.

\bibitem{feichtenhofer2015dynamically}
C.~Feichtenhofer, A.~Pinz, and R.~P. Wildes.
\newblock Dynamically encoded actions based on spacetime saliency.
\newblock In {\em CVPR}, 2015.

\bibitem{fernando2015modeling}
B.~Fernando, E.~Gavves, J.~M. Oramas, A.~Ghodrati, and T.~Tuytelaars.
\newblock Modeling video evolution for action recognition.
\newblock In {\em CVPR}, 2015.

\bibitem{jegou2010aggregating}
H.~J{\'e}gou, M.~Douze, C.~Schmid, and P.~P{\'e}rez.
\newblock Aggregating local descriptors into a compact image representation.
\newblock In {\em CVPR}, 2010.

\bibitem{kantorov2014efficient}
V.~Kantorov and I.~Laptev.
\newblock Efficient feature extraction, encoding and classification for action
  recognition.
\newblock In {\em CVPR}, 2014.

\bibitem{kuehne2011hmdb}
H.~Kuehne, H.~Jhuang, E.~Garrote, T.~Poggio, and T.~Serre.
\newblock Hmdb: a large video database for human motion recognition.
\newblock In {\em ICCV}, 2011.

\bibitem{lan2015beyond}
Z.~Lan, M.~Lin, X.~Li, A.~G. Hauptmann, and B.~Raj.
\newblock Beyond gaussian pyramid: Multi-skip feature stacking for action
  recognition.
\newblock In {\em CVPR}, 2015.

\bibitem{lucas1981iterative}
B.~D. Lucas and T.~Kanade.
\newblock An iterative image registration technique with an application to
  stereo vision.
\newblock {\em IJCAI}, 81:674--679, 1981.

\bibitem{peng2014action}
X.~Peng, C.~Zou, Y.~Qiao, and Q.~Peng.
\newblock Action recognition with stacked fisher vectors.
\newblock In {\em ECCV}. 2014.

\bibitem{perronnin2007fisher}
F.~Perronnin and C.~Dance.
\newblock Fisher kernels on visual vocabularies for image categorization.
\newblock In {\em CVPR}, 2007.

\bibitem{simonyan2014two}
K.~Simonyan and A.~Zisserman.
\newblock Two-stream convolutional networks for action recognition in videos.
\newblock In {\em NIPS}, 2014.

\bibitem{Simonyan14c}
K.~Simonyan and A.~Zisserman.
\newblock Very deep convolutional networks for large-scale image recognition.
\newblock In {\em ICLR}, 2015.

\bibitem{soomro2012ucf101}
K.~Soomro, A.~R. Zamir, and M.~Shah.
\newblock Ucf101: A dataset of 101 human actions classes from videos in the
  wild.
\newblock {\em arXiv:1212.0402}, 2012.

\bibitem{sun2015human}
L.~Sun, K.~Jia, D.-Y. Yeung, and B.~E. Shi.
\newblock Human action recognition using factorized spatio-temporal
  convolutional networks.
\newblock In {\em ICCV}, 2015.

\bibitem{tran2014c3d}
D.~Tran, L.~Bourdev, R.~Fergus, L.~Torresani, and M.~Paluri.
\newblock Learning spatiotemporal features with 3d convolutional networks.
\newblock In {\em ICCV}, 2015.

\bibitem{wang2011action}
H.~Wang, A.~Kl{\"a}ser, C.~Schmid, and C.-L. Liu.
\newblock Action recognition by dense trajectories.
\newblock In {\em CVPR}, 2011.

\bibitem{wang2013action}
H.~Wang and C.~Schmid.
\newblock Action recognition with improved trajectories.
\newblock In {\em ICCV}, 2013.

\bibitem{wang2015action}
L.~Wang, Y.~Qiao, and X.~Tang.
\newblock Action recognition with trajectory-pooled deep-convolutional
  descriptors.
\newblock In {\em CVPR}, 2015.

\bibitem{wang2015towards}
L.~Wang, Y.~Xiong, Z.~Wang, and Y.~Qiao.
\newblock Towards good practices for very deep two-stream convnets.
\newblock {\em arXiv:1507.02159}, 2015.

\bibitem{wu2015modeling}
Z.~Wu, X.~Wang, Y.-G. Jiang, H.~Ye, and X.~Xue.
\newblock Modeling spatial-temporal clues in a hybrid deep learning framework
  for video classification.
\newblock In {\em ACMMM}, 2015.

\bibitem{xu2015discriminative}
Z.~Xu, Y.~Yang, and A.~G. Hauptmann.
\newblock A discriminative {CNN} video representation for event detection.
\newblock In {\em CVPR}, 2015.

\bibitem{yue2015beyond}
J.~Yue-Hei~Ng, M.~Hausknecht, S.~Vijayanarasimhan, O.~Vinyals, R.~Monga, and
  G.~Toderici.
\newblock Beyond short snippets: Deep networks for video classification.
\newblock In {\em CVPR}, 2015.

\end{thebibliography}

\clearpage
\section*{Supplemental Material}

\setcounter{section}{0}
\renewcommand{\thesection}{\Alph{section}}

\section{Parameter on UCF101 split1}

Following previous work that encodes CNN-based local descriptors \cite{xu2015discriminative}, we first evaluate dimension reduction. Then, we explore the number of clusters for encoding.

\subsection{FV}
We first evaluate descriptor dimensions after compression by PCA with a fixed number of gaussian mixtures $K=256$. Table \ref{tab:pca_fv} shows that 64-D achieves the best performance on all methods. Thus, we employ 64-D for FV.

Next we evaluate number of gaussian mixtures $K$. Table \ref{tab:gmm_fv} shows that $K=128$ achieves the best result on TDD and TDD + CPD, while $K=256$ performs the best on CPD. However, $K=128$ on CPD shows comparable performance to $K=256$. Thus, we fixed $K=128$ both on TDD and CPD in this paper.

\begin{table}[ht]
\begin{center}
     \caption{Impact of TDD and CPD dimensions after compression with fixed $K=256$ in FV.}
  \begin{tabular}{|c|c|c|c|c|}
\hline
Dimensions				&32-D	&{\bf 64-D}	&128-D	&256-D	\\ \hline
TDD						&90.3\%	&{\bf 90.4\%}		&90.4\%	&90.2\%	\\ \hline
CPD						&90.0\%	&{\bf 90.6\%}		&90.4\%	&90.2\%	\\ \hline
TDD+CPD				&90.2\%	&{\bf 90.8\%}		&90.5\%	&90.5\%	\\ \hline
  \end{tabular}
       \label{tab:pca_fv}
   \end{center}
\end{table}

\begin{table}[ht]
\begin{center}
     \caption{Impact of the number of gaussian mixture $K$ with fixed PCA dimensions of 64-D in FV.}
  \begin{tabular}{|c|c|c|c|c|}
\hline
Clusters					&$K=32$	&$K=64$	&${\bf K=128}$	&$K=256$		\\ \hline
TDD						&89.5\%	&90.4\%	&{\bf 90.7\%}	& 90.4\%	\\ \hline
CPD						&89.9\%	&90.2\% 	&90.4\% 	& 90.6\%	\\ \hline
TDD+CPD				&90.2\%	&90.6\% 	&{\bf 90.8\%}& 90.8\%	\\ \hline
  \end{tabular}
       \label{tab:gmm_fv}
   \end{center}
\end{table}

\newpage

\subsection{VLAD}
We also evaluate dimensions and number of clusters in VLAD. Table \ref{tab:pca_vlad} shows that 128-D achieves the best performance on CPD and TDD + CPD. Although 128-D is not the best on TDD, it achieves comparable performance. Thus, we employ 128-D for VLAD.
Next we evaluate number of k-means clusters $K$. Table \ref{tab:kmeans_vlad} shows its result. We can see the best $K$ is 64 on CPD and TDD + CPD. $K=64$ also achieves almost the same result as the best one, $K=32$ on TDD. Thus, we fixed $K=64$ both on TDD and CPD in this paper.

\begin{table}[ht]
\begin{center}
     \caption{Impact of the TDD and CPD dimensions after compression with fixed $K=256$ in VLAD.}
  \begin{tabular}{|c|c|c|c|c|}
\hline
Dimensions				&32-D	&64-D	&{\bf 128-D}	&256-D	\\ \hline
TDD						&91.2\%	&91.1\%	&90.9\%		& 91.1\%	\\ \hline
CPD						&90.7\%	&91.2\% 	&{\bf 91.5\%}	& 91.5\%	\\ \hline
TDD+CPD				&91.5\%	&91.2\%	&{\bf 91.5\%} 	& 91.4\%	\\ \hline
  \end{tabular}
     \label{tab:pca_vlad}
   \end{center}
\end{table}

\begin{table}[ht]
\begin{center}
     \caption{Impact of the number of k-means clusters $K$ with fixed PCA dimensions of 128-D in VLAD.}
  \begin{tabular}{|c|c|c|c|c|}
\hline
Clusters					&$K=32$		&${\bf K=64}$		& $K= 128$	&$K=256$		\\ \hline 
TDD						&{\bf 91.6\%}	&91.5\%		&91.3\%		&90.9\%	\\ \hline
CPD						&90.3\%		&{\bf 91.6\%}		&91.3\% 		&91.5\%	\\ \hline
TDD+CPD				&91.6\%		&{\bf 92.0\%}		&91.5\%		&91.5\%	\\ \hline
  \end{tabular}
     \label{tab:kmeans_vlad}
   \end{center}
\end{table}

\end{document}